# Overcoming Hierarchical Difficulty by Hill-Climbing the Building Block Structure

David Iclănzan
Sapientia Hungarian University of Transylvania
547367, Corunca, Şoseaua Sighişoarei 1C
Tg-Mureş, ROMANIA
david.iclanzan@gmail.com

Dan Dumitrescu
Babeş-Bolyai University
400084, Str. Mihail Kogălniceanu nr. 1
Cluj-Napoca, ROMANIA
ddumitr@cs.ubbcluj.ro

## ABSTRACT
The Building Block Hypothesis suggests that Genetic Algorithms (GAs) are well-suited for hierarchical problems, where efficient solving requires proper problem decomposition and assembly of solution from sub-solution with strong non-linear interdependencies. The paper proposes a hill-climber operating over the building block (BB) space that can efficiently address hierarchical problems. The new Building Block Hill-Climber (BBHC) uses past hill-climb experience to extract BB information and adapts its neighborhood structure accordingly. The perpetual adaptation of the neighborhood structure allows the method to climb the hierarchical structure solving successively the hierarchical levels. It is expected that for fully non deceptive hierarchical BB structures the BBHC can solve hierarchical problems in linearithmic time. Empirical results confirm that the proposed method scales almost linearly with the problem size thus clearly outperforms population based recombinative methods.

## Categories and Subject Descriptors
G.1.6 [**Mathematics of Computing**]: Global Optimization–Analyze; I.2.8 [**Artificial Intelligence**]: Problem Solving, Control Methods, and Search

## 1. INTRODUCTION
One of the most important research goal regarding Evolutionary Algorithms (EAs) is to understand the class of problems for which these algorithms are most suited. Despite the major work in this field it is still unclear how an EA explores a search space and on what fitness landscapes will a particular EA outperform other optimizers such as hill-climbers.

The traditional GA theory is pillared on the Building Block Hypothesis (BBH) which states that GAs work by discovering, emphasizing and recombining low order schemata in high-quality strings in a strongly parallel manner [1, 2].

Albeit being widely used to justify claims about EAs, the BBH remains controversial [3, 4]. While there are situations where the BBH provides a good explanation about how a GA works there are also cases, like GA with uniform crossover performing well on certain test suits, where BBH in the current form does not provide much useful insight.

In the early 90's a systematic program was initiated by Mitchell et al [5] to address these issues concerning the fundamentals of GAs. Their strategy was to find a set of features that are of particular relevance to GAs and test the performance of GAs on landscapes containing those features. It was recognized that major tenets behind the BBH are the notion of problem decomposition and the assembly of solutions from sub-solutions. Subsequently they constructed a set of functions that clearly emphasize a gross-scale BB structure with low-order BBs that recombine to higher-order ones. These functions were expected to lay out a "royal road" for GAs while hill-climbers were anticipated to perform poorly as a large number of positions must be optimized simultaneously to discover higher-order BBs. To much of a surprise both expectation were refuted: on these test suits GAs performed worse than expected due to the hitchhiking phenomena while a Random Mutation Hill-Climber which accepts states with equal objective function value greatly outperformed GAs.

Later developments proposed NK landscapes [6] and the expanded function method [7] which presented non-separable components on a single level.

Inspired by the fact that many real-world systems are hierarchical, Watson et al. [8] proposed a class of hierarchically decomposable functions which present a strong non-linear hierarchical BB interdependency. This class of function is very hard for mutation based hill-climbers as the Hamming distance between local optima and global optima is very large. It is considered that this class of functions exemplifies those problems for which GAs are well-suited.

The objective of this paper is to develop a hill-climber that can solve hierarchical problems. The proposed method operates on BBs rather than bits and uses past experience to learn linkages and for adapting its neighborhood structure. The neighborhood structure adaptation allows the hill-climber to hierarchically decompose the problems, revealing the hierarchical levels one after the other; when it finishes the problem structure is delivered in an explicit manner that is transparent to human researchers.

The following section summarizes the characteristics of hierarchically decomposable problems. Section 3 revisits the notion of hill-climbing; introduces and informally describes the concept of BB hill-climbing. The proposed method is presented in details in Section 4. Section 5 presents empirical results of the proposed method. Finally, the paper is concluded in Section 6 by discussion and some future works.

## 2. HIERARCHICALLY DECOMPOSABLE FUNCTIONS

Although having a gross-scale BB structure, hierarchical problems are hard to solve without proper problem decomposition as the blocks from these functions are not separable.

The fundamental of hierarchically decomposable problems is that there is always more than one way to solve a (sub-) problem [8] leading to the separation of BBs "fitness" i.e. contribution to the objective function, from their meaning. This conceptual separation induces the non-linear dependencies between BBs: providing the same objective function contribution, a BB might be completely suited for one context whilst completely wrong for another one. Thus the "fitness" of a BB can be misleading if it is incompatible with its context. However, the contribution of the BBs indicate how can the dimensionality of the problem be reduced by expressing one block in a lower level as one variable in the upper level.

Hierarchical problems are very hard for mutation based hill-climbers as they exhibit a fractal like structure in the Hamming space with many local optima [9]. This bit-wise landscape is fully deceptive; the better is a local optimum the further away is from the global ones. At the same time the problem can be solved quite easily in the BB or "crossover" space, where the "block-wise landscape" is fully non-deceptive [8]. The forming of higher order BBs from lower level ones reduces the problem dimensionality. If a proper niching is applied and the promising sub-solutions are kept until the method advances to upper levels where a correct decision can be made, the hierarchical difficulty can be overcome.

Several methods are known which optimize problems with random linkage by hierarchical decomposition. They can naturally be divided in two classes according to how the decomposition information is stored. In Pelikan and Goldberg's approach [10], the Hierarchical Bayesian Optimizer stores the decomposition information implicitly in a Bayesian network. Another method, the DSMGA++, recently proposed by Yu and Goldberg [11] which decompose problems by using dependency structure matrix clustering techniques and stores the decomposition information *explicitly* thus being able to deliver the problem structure in a comprehensible manner for humans.

### 2.1 Hierarchical Problems

In this paper three hierarchical test functions are used: the hierarchical IFF [8], the hierarchical XOR [12] and the hierarchical trap function [10]. These problems are defined on binary strings of the form $x \in \{0,1\}^{k^p}$, where $k$ is the number of sub-blocks in a block, and $p$ is the number of hierarchical levels. The meaning of sub-blocks is separated from their fitness by the means of a boolean function $h(b)$ which determines if the sub-block $b$ is *valid* in the current context or not. In the shuffled version of these problems the tight linkage is disrupted by randomly reordering the bits. The functions with their particularities are detailed as follows.

#### 2.1.1 The Hierarchical If and Only If (hIFF)

The hIFF has $k = 2$ and it is provided by the if and only if relation, or equality. Let $L = x_1 x_2 ... x_{2^{p-1}}$ be the first half of the binary string $x$ and $R = x_{2^{p-1}+1} x_{2^{p-1}+2} ... x_{2^p}$ the second one. Then $h$ is defined as:

$$h_{iff}(x) = \begin{cases} 1, & if\ p = 0 \\ 1, & if\ h_{iff}(L) = 1, h_{iff}(R) = 1\ and\ L = R \\ 0, & otherwise. \end{cases}$$

Based on $h_{IFF}$ the value of hIFF is defined recursively:

$$H_{iff}(x) = H_{iff}(L) + H_{iff}(R) + \begin{cases} length(x), if\ h_{iff}(x) = 1 \\ 0,\ otherwise \end{cases}$$

At each level $p > 0$ the $H_{iff}(x)$ function rewards a block if and only if the interpretation of the two composing sub-blocks are both either 0 or 1. Otherwise the contribution is zero.

hIFF has two global optima: strings formed only by 0's or only by 1's. At the lowest level the problem has $2^{l/2}$ local optima where $l$ is the problem size.

#### 2.1.2 The Hierarchical XOR (hXOR)

The definition of hXOR is analogous with the hIFF, having only a modification in the validation function $h$, where instead of equality we do a complement check.

$$h_{iff}(x) = \begin{cases} 1, & if\ p = 0 \\ 1, & if\ h_{iff}(L) = 1, h_{iff}(R) = 1\ and\ L = \overline{R} \\ 0, & otherwise. \end{cases}$$

The $\overline{R}$ stands for the bitwise negation of $R$.

The two global optima of hXOR are composed by half zeros and half ones. Having the same problem structure it is expected that an algorithm which apply problem decomposition to perform equally well on both problems. This is not always the case as some methods may be biased to replicate particular alleles, solving the hIFF in an easier manner.

#### 2.1.3 The Hierarchical Trap function (hTrap)

The underlying structure of the hTrap is a balanced $k$-ary tree where $k \geq 3$. Blocks from lower level are interpreted by a *mapping function* similar to the one from the hIFF: a block of all 0's and 1's is mapped to 0 and 1 respectively, and everything else is interpreted as '-' or *null*.

The contribution function is a trap function of unitation (its value depends only on the numbers of 1's in the input string) of order $k$, based on two parameters $f_{high}$ and $f_{low}$ which defines the degree of deception.

Let $u$ be the unitary of the input string. Then the trap function is defined as:

$$trap_k(u) = \begin{cases} f_{high}, & if\ u = k \\ f_{low} \times \dfrac{k-1-u}{k-1}, & otherwise \end{cases}$$

If any position in the input string is *null* ('-') then the contribution is zero.

In this paper we use hTrap function based on $k = 3$, $f_{high}$ and $f_{low}$ set to 1 for all except the highest level. The decision between competing BBs can be carried out only on the highest level, where $f_{high} = 1$ and $f_{low} = 0.9$.

## 3. HILL-CLIMBING IN THE BUILDING BLOCK SPACE

Hill-climbing is used widely in artificial intelligence fields, for quickly reaching a goal state from a starting position. The hill-climbers are usually the fastest methods but they get trapped in local optima on deceptive landscapes. The current section revisits the notion of hill-climbing and neighborhood structure and introduces the idea of BB hill-climber that can solve hierarchical problems by exploiting the BB structure and adapting its neighborhood structure online.

### 3.1 Hill-climbers and neighborhood structure

Hill-climbing is an optimization technique that starts from some initial solution and iteratively tries to replace the current solution by a better one from an appropriately defined neighborhood of the current state. In *simple* or *first improvement hill-climbing*, the first better solution is chosen whereas in *steepest ascent* or *best improvement hill-climbing* all successors are compared and the best solution is chosen.

Random-restart hill-climbing (Figure 1) simply runs an outer loop over hill-climbing. Each step of the outer loop chooses a random initial state $s$ to start hill-climbing. The best solution encountered is kept.

The neighborhood structure is defined as follows:

*Let $S$ be the solution space. Then the neighborhood structure is a function $N : S \to 2^S$ that assigns to every $s \in S$ a set of neighbors $N(s) \subseteq S$. $N(s)$ is called the neighborhood of $s$.*

Usually hill-climbers described in the literature use bit-flipping for replacing the current state. This implies a neighborhood structure which contains strings that are relatively

1. Generate a random state $s$;
2. Hill-climb from $s$;
3. If the resulted state is better then the best states seen so far, keep the new state.
4. if termination condition not met goto 1;

**Figure 1. Outline of the random restart hill-climber.**

close in Hamming distance to the original state, making those methods unsuited for solving hierarchical problems where local optima and global optima are distant in Hamming space. But $N$ can be any function; the main idea of the paper is to build a hill-climber which takes into account the BB structure of the problems and defines its neighborhood structure accordingly.

### 3.2 New Approach: Building Block Hill-climbing

As already indicated in Section 2, hierarchical problems are fully deceptive in Hamming space and fully non deceptive in the BB space. The problem representation together with the neighborhood structure defines the search landscape. With an appropriate neighborhood structure, which operates on BBs, the search problem can be transferred from Hamming space to a very nice, fully non deceptive search landscape which should be easy to hill-climb.

In the proposed model the individual is represented as a sequence of BBs:

$s = (b_1, b_2 ... b_n)$ where $n$ is the number of BBs.

Each BB $b_i$ can represent multiple configurations: $V_i = \{v \mid v \in \{0,1\}^{l_i}\}$ where $l_i$ is the length of $b_i$. In this way competing schemata can be kept. For example in the case of the hIFF a BB of length 2 may have two valuable configurations: $V = \{(0,0),(1,1)\}$.

The neighborhood structure of a state $s$ is composed by all possible combinations of BB configurations:

$$N(s) = \{z \mid z \in V_1 \times V_2 \times ... \times V_n\}$$

Instead of flipping bits as in classical mutation based hill-climbers, the proposed method hill-climbs the BB structure by choosing the best configuration of every BB in a greedy manner. The hill-climb on the BB structure will get the state $s$ to the nearest local optima.

While EAs exploit BB structure by probabilistic recombination, this approach applies a systematic combination and analysis of BBs.

## 3.3 Online adaptation of the neighborhood structure from past experience

It is important for a GA to conserve BBs under crossover. Theoretical studies denote that a GA that uses crossover which does not disrupts the BB structure holds many advantages over simple GA [13]. To achieve this goal linkage learning is applied and the solution representation is evolved along with the population, during the search process.

Similarly, in order to be able to hill-climb the BB landscape, the BB structure of the problem must be learned and the representation of the individual must be evolved to reflect the current BB knowledge. The changing of representation implies the adaptation of the neighborhood structure which is the key to conquer hierarchical problems: by exploring the neighborhood of the current BB configuration the next level of BB can be detected.

In order to be able to identify linkages we enhance the hill-climber with a memory where hill-climbing results are stored. Evolutionary Algorithms with linkage learning mechanism extracts the BB information from the population. Similar techniques can be applied to devise BB structures from past experience stored in the memory. However, the solutions stored in the memory offer an important advantage over populations: they are noise free. While individuals from populations may have parts where good schemata's have not yet been expressed or have BBs slightly altered by mutation, the solution stored in memory are always exact local optima. In the case of fully non deceptive BB landscapes the systematic exploration of BB configurations guarantees that in the close neighborhood of these states there are no better solutions.

The details of linkage learning mechanism and BB construction are detailed in the next section.

## 4. THE BUILDING BLOCK HILL-CLIMBER

BBHC involves three main steps: (i) hill-climbing the search space according to a BB neighborhood structure; (ii) local optimum obtained in (i) is used to detect linkages and extract BB information; (iii) the BB configuration and implicitly the neighborhood structure are updated. The section describes the framework of BBHC and indicates how the three steps are implemented.

### 4.1 The BBHC Framework

Figure 2 depicts the two main phases of the BBHC optimization. The first one refers to the accumulation of search experience provided by the repeated hill-climbing. The second phase concerns the exploitation of search experience by linkage learning and BB structure updating.

The input of the second phase is the search experience stored in memory. Dependencies are detected and the output consists of updated BB structure which enables the first phase to combine new BBs. In hierarchical problems modeled after the suggestions of BBH the assembling of lower level BBs leads to the development of higher order BBs. Thus the sequence of phases can effectively overcome hierarchical levels successively by discovering and incorporating BB knowledge in the search process.

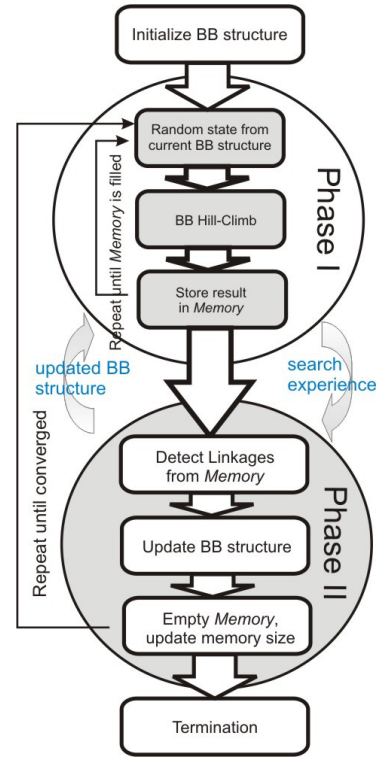

**Figure 2. The framework of BBHC with the two main phases: accumulation and exploitation of search experience.**

### 4.2 The BB Hill-Climbing

BB hill-climbing is rather straightforward - instead of flipping bits, the search focuses on the best local BB configuration. Each BB is processed systematically by testing its configurations and selecting the one which provides the highest (or lowest in the case of minimization) objective function value. While the best configuration of a particular BB is searched, the configurations of the other BBs are hold still. The BB hill-climbing technique is outlined in Figure 4.

As an example let us consider the current BB structure (denoted by $BB(s)$) of an 8-bit state, described by three BBs: $BB(s) = (b_1, b_2, b_3)$. The BBs describe the loci of the state in the following way (Figure 3):

$b_1 = \{locus\, 2,\ locus\, 6\}$, $b_2 = \{locus\, 3,\ locus\, 5\}$ and $b_3 = \{locus\, 1,\ locus\, 4,\ locus\, 7,\ locus\, 8\}$.

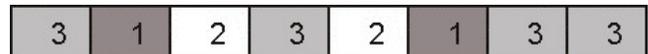

**Figure 3. BB configuration of the 8-bit state.**

1. Choose randomly an unprocessed building block $b_i$ from $s$;
2. Choose randomly an unprocessed building block configuration $v_j \in V_i$;
3. Set $v(b_i)$ in $s$ to $v_j$;
4. If the change results in a decrease of the objective function undo the change;
5. If there exists unprocessed building block configuration of $b_i$ goto 2;
6. If there exists unprocessed building block from $s$ goto 1;

**Figure 4. The pseudo code of the building block hill-climbing.**

The configurations of the BBs are: $V_1 = \{00, 11\}$, $V_2 = \{00, 01, 11, 10\}$ and $V_3 = \{0000, 1111\}$. A BB $b_i$ will always have a configuration from $V_i$ and the index of this value is denoted by $v(b_i)$.

The state $s = (2, 3, 1)$ expresses the bit configuration obtained from the second configuration of $b_1$, the third configuration of $b_2$ and finally the first configuration of $b_3$: $decode(s) = 01101100$.

In what follows it is explained how $b_2$ is hill-climbed starting from the state $s = (2, 3, 1)$. All configurations of $b_2$ are analyzed in the current context where $v(b_1) = 2$ and $v(b_3) = 1$ are fixed. The neighboring states of $s$ are $s_1 = (2, \mathbf{1}, 1)$, $s_2 = (2, \mathbf{2}, 1)$, $s_3 = (2, \mathbf{3}, 1)$, $s_4 = (2, \mathbf{4}, 1)$, corresponding to the bit configurations $decode(s_1) = 01\mathbf{000}100$, $decode(s_2) = 01\mathbf{001}100$, $decode(s_3) = 01\mathbf{101}100$, $decode(s_4) = 01\mathbf{100}100$. The most suitable configuration for $b_2$ is chosen with respect to the objective function and the original state is updated accordingly. For instance, if the first configuration provides the best objective function value then the state becomes $s = (2, \mathbf{1}, 1)$. The algorithm carries on by hill-climbing another BB in the same manner from the new state $s$.

After all BBs are processed the result is stored in the memory and the process is repeated starting from a new random state until the memory is full. Next, linkages from the memory are detected and the BB structure is updated

## 4.3 Linkage Detection and BB structure Update

Several techniques for detecting gene dependency from a population are presented in the literature [14, 11]. Similar techniques can also be used to detect the linkage from the states

1. Choose randomly a building block $b_i$ from $s$ which has not yet been clustered;
2. Let $L$ be the set of building blocks whose configuration from the memory are mapped bijectively to $b_i$;
3. If $L$ is empty update the possible configurations $V_i$ to the configurations encountered in the memory;
4. If $L$ is not empty form a new building block $new\_b = b_i \bigcup L$ by setting the loci it's define to the union of loci from $b_i$ and the building blocks from $L$. Also set the possible values $V_{new\_b}$ to all distinct configuration encountered, on the position defined by the $new\_b$, operating on the binary representation of states from the memory.
5. Set $b_i \bigcup L$ as clustered;
6. If there exists building blocks which have not been clustered goto 1;

**Figure 5. The linkage detection and new BB forming algorithm.**

stored in the memory. . Due to the fact that the hierarchical problems studied have a very nice and easy structure in the BB space a very simple method for linkage detection is considered. The process is facilitated by the advantage of having noise free states stored in memory. Probabilistic methods may involve many individuals which do not express an accurate BB configuration whilst the states from the memory are always exact local optima.

The clustering of loci in new BBs is done by searching for bijective mappings. For a given block $b_i$, all BBs $b_j$ are linked if distinct configurations of $b_i$ map to distinct configurations of $b_j$. The configurations of $b_i$ that can be found in the memory represent the domain while the configurations of $b_j$ from the memory are the codomain. Thus we say that $b_i$ and $b_j$ are linked if there is a one-to-one correspondence between their configurations i.e. for any particular value $v_i$ from the domain there exist a unique configuration $v_j \in V_j$ satisfying the condition: for every state from the memory if $v(b_i) = v_i$ then $v(b_j) = v_j$.

Due to the transitivity property of bijective mappings (functions) all relevant BBs are discovered simultaneously. The linkage detection algorithm is presented in Figure 5. Harder problems (exhibiting overlapping BB structure for example), may require a more sophisticated linkage learning method.

All BBs linked together by a bijective mapping will form a new BB which replaces the linked loci in the BB structure. The possible configurations of the new BBs are extracted from the binary representation of states from the memory. All distinct configurations from the positions defined by the composing BBs

1. Generate a random state $s$ from the current BB structure;
2. BB cill-climb from $s$ and store the result in memory;
3. If the resulted state is better then the best states seen so far, keep the new state;
4. If the memory is not filled up goto 1;
5. Learn linkage from memory and update the BB configuration according to the detected linkages;
6. Empty memory;
7. If necessary update *MemorySize*;
8. If termination condition not met goto 1;

**Figure 6. Outline of the hill-climbing enhanced with memory and linkage learning. In steps 1-4 we accumulate the search experience (phase 1) which is exploited in steps 5-8 (phase 2).**

are taken into account. If a BB can not be linked with any other BB it keeps its original place and only its possible configurations are updated in the same manner as the new BBs.

Revisiting the example from previous section, let us suppose that the memory contains the following hill-climbed states for the BB configuration $s = (b_1, b_2, b_3)$: $M$ = {(1,1,1), (2,3,2), (1,1,1), (2,3,1)} corresponding to the binary representation $M_{bin}$ = {(00000000), (11111111), (00000000), (01101100)}. Next the linkages are detected starting from the first locus. $b_1$ can be linked with $b_2$ as its every distinct configuration can be mapped to a distinct configuration of $b_2$: $\langle v(b_1)=1 \rangle \to \langle v(b_2)=1 \rangle$ and $\langle v(b_1)=2 \rangle \to \langle v(b_2)=3 \rangle$. Although $v(b_1)=1$ maps to $v(b_3)=1$ we do not have a linkage as for $v(b_1)=2$ there are two distinct values (1 and 2) for $b_3$. Even if supposedly all the configurations from the memory for $b_3$ are 1, we still can not establish a linkage between $b_1$ and $b_3$ because there is no univocal correspondence (mapping): for both $v(b_1)=1$ and $v(b_1)=2$, $v(b_3)$ would yield 1.

From $b_1$ and $b_2$ a new BB denoted by $b_4$ is formed. The positions which it defines in the state are obtained by the union of loci defined by $b_1$ and $b_2$: $b_4 = \{locus\,2, locus\,3, locus\,5, locus\,6\}$. The possible configurations are defined by the distinct configurations found in the memory on the loci defined by the union of $b_1$ and $b_2$: $V_4 = \{0000, 1111\}$. The BB structure is updated according to detected linkages, $b_1$ and $b_2$ being replaced by $b_4$. The BB structure is set to $s = (b_3, b_4)$ and the hill-climbing is resumed in the combinative neighborhood of these two BBs, which will hopefully lead to new linkages.

As linkages are successively detected and bigger and bigger BBs are formed, the dimensionality of the problem is reduced. Lower dimensionality translates to less search experience needed to detect linkages. Thus with the reduction of the dimensionality the size of the memory - the number of BB hill-climbs performed before we proceed to linkage learning – can also be reduced. For fully non deceptive problems in the BB space a number of solutions proportional with the logarithm of the state length should be enough to successfully detect linkages. In this paper we use the formula $memory\_size = c + \lfloor \log(length(s)) \rfloor$ where $c$ is a constant.

Proposed model can be summarized by the algorithm presented in Figure 6.

## 5. RESULTS

We tested the scalability of BBHC on 128-bit, 256-bit, 512-bit and 1024-bit shuffled hIFF and hXOR problems, respectively on 81-bit, 243-bit and 729-bit shuffled hTrap problem. Lower problem sizes were not addressed as they may be too easy to solve; any conclusion from them may be misleading.

The size of the memory was set to $8 + \lfloor \log_2(length(s)) \rfloor$ on hIFF and hXOR respectively to $18 + \lfloor \log_3(length(s)) \rfloor$ on hTrap.

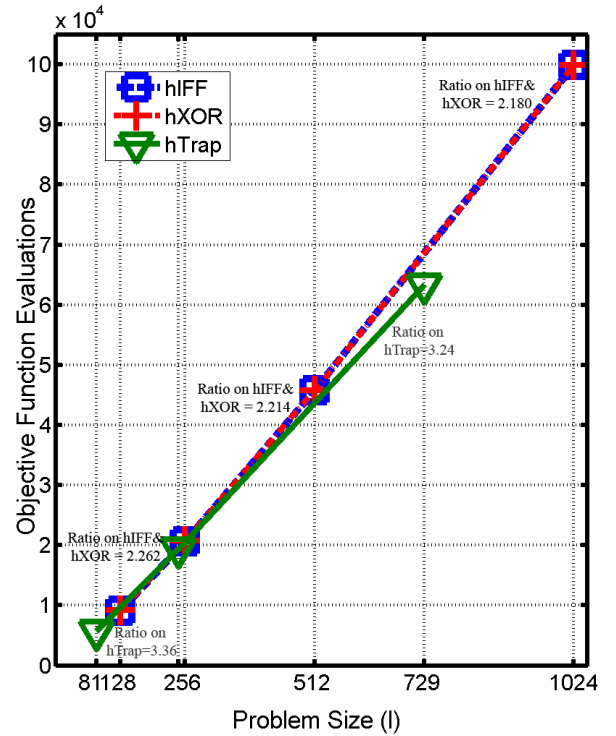

**Figure 7. Arithmetic scaling of BBHC. Preliminary scalability test of BBHC on hIFF, hXOR and hTrap. The number of function evaluations scales up almost linearly. The ratio between neighborhood points is decreasing towards 2 as the problem sizes are doubled in the case of hIFF and hXOR and towards 3 in the case of the hTRap function.**

The arithmetic scaling results with the ratio between neighborhood points is presented in Figure 7. The proposed method scales up almost linearly with the problem size on the test suits with the slope between neighbor points decreasing towards 2 as the problem size doubles in the case of hIFF and hXOR and towards 3 on hTrap as problem size is tripled. The almost linear scaling is the direct result of the fact that each level of the hierarchical problems is solved by the BBHC with $O(L\log(L))$ complexity, where $L$ is the size of the level. The slopes of hIFF and hXOR coincide due to the identical problem structure.

The experimental results are approximated with functions of the form $f(x) = ax^b \log(x)$ where $a$ and $b$ are determined by the least square error method.

In Figure 8 the log-log scaled plot of the test results and of the approximation functions are shown. The approximated number of objective function evaluations scales as $O(l^{0.97}\log(l))$ on hIFF and hXOR and $O(l^{0.91}\log(l))$ on hTrap, where $l$ is the problem size. The approximations are very close to the expected linearithmic $O(l\log(l))$ time. The best results reported till now scale up sub-quadratically with an expected lower bound of $O(l^{1.5}\log(l))$ [15].

One of the best known optimizers that operate via hierarchical decomposition, the hBOA, with hand tuned parameters solves the 256-bit shuffled hIFF in approximately 88000 function evaluations. The BBHC performance on the same test suit is 20666, approximately four times quicker than the hBOA. Due to the almost linear scaling the BBHC is able to solve the 512-bit version of the same problem approximately twice as fast as the hBOA does the 256-bit one, requiring only 45793 function evaluations!

Similarly to other methods like the DSMGA++, the BBHC uses explicit chunking mechanism enabling the method to deliver the problem structure. While DSGMA++ and other stochastic methods have to fight the sampling errors which sometimes induce imperfections, the BBHC was able to detect the perfect problem structure in all runs, due to its more systematic and deterministic approach. The enhanced capability of BBHC to capture the problem structure is also revealed by the fact that hIFF and hXOR are solved approximately in the same number of steps as their underlying BB structures (a balanced binary tree) coincide. However for the DSGMA++ the time needed to optimize the two problems differs significantly, being $O(l^{1.84}\log(l))$ for the hIFF and $O(l^{1.96}\log(l))$ on hXOR.

On hIFF and hXOR where there are two global optima, the multiple runs of the BBHC showed an unbiased behavior, finding in almost half-half proportion both solutions.

A final remark concerns the stability of BBHC: the highest standard deviation encountered is 210 while other methods deal with standard deviations of much higher magnitude on the same test suites.

## 6. CONCLUSIONS AND FUTURE WORKS

The concept of BB hill-climber (BBHC) a generic method for solving problems via hierarchical decomposition is proposed. The BBHC operates in the BB space where it combines BBs in a systematic and exhaustive manner. Past hill-climbing experience is used to learn the underlying BB structure of the search space expressed by linkages. The continuous update of the BB representation of the individual results implicitly in the adaptation of the neighborhood structure to the combinative neighborhood of the current BB representation. In hierarchical problems - where BBH holds - moving the search to the combinative vicinity of the current BB representation facilitates the discovery of new BBs, as BBH imply that low-order BBs can be combined to form higher-order ones.

An important aspect of the proposed method is that similar to DSMGA++, BBHC delivers the problem structure in a form comprehensible to humans. Gaining knowledge about the hidden, complex problem structure can be very useful in many real-world applications.

BBHC clearly outperforms population based recombinative methods on different issues of hierarchical problems. So far hBOA proved the greatest ability to solve hierarchical problems with random linkage. Nevertheless, the proposed method solves the 512-bit shuffled version of hIFF and hXOR approximately twice as fast then the hBOA solves the smaller version of 256-bit!

Preliminary scalability test of the proposed method indicates that BBHC holds not only a quantitative advantage over other methods but also a qualitative one too: it scales linearithmic with the problem size.

This result reposes questions on the fundamentals of EAs as we now must again look for different kind of problems to exemplify the utility of the nature inspired algorithms. Also the suggestion of BBH regarding the use of population is put at question: according to the BBH population is needed to combine BBs in a highly parallel manner. However the result of the paper shows that a systematic exploration of BBs combination by a hill-climber can be more efficient.

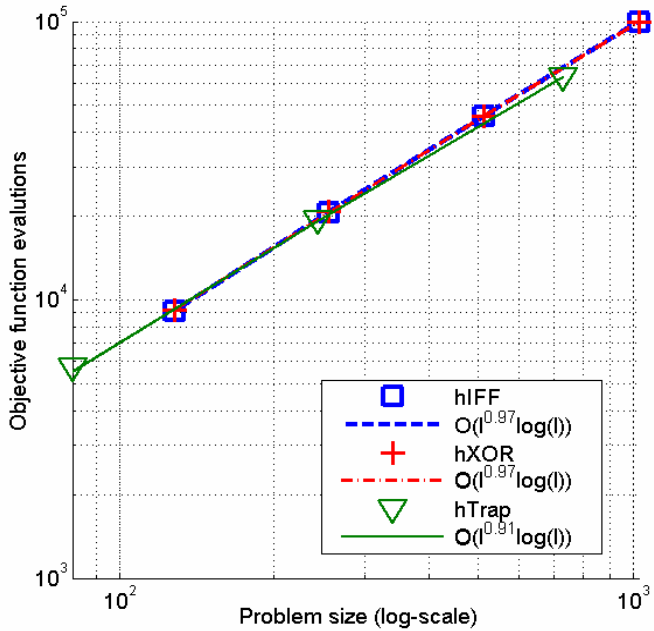

**Figure 8. The number of function evaluations of BBHC approximately scales as $O(l^{0.97}\log(l))$ on hIFF and hXOR and $O(l^{0.91}\log(l))$ on hTrap, where $l$ is the problem size.**

The No Free Lunch Theorem [16, 17] guarantees that there exits problems where GAs outperform other methods. The main issue regards relevance: how much if any this class of function is related to real-world problems?

We think that the task of finding problems for which GAs are well suited must be approached from direction that so far got little or no attention in the literature due to doubtful expectances regarding the role of different features of GAs. The results of the paper suggest that if a problem has a nice structure, even if "hidden" like the BB space, a proper hill-climber can outperform population based recombinative methods, *without requiring extra domain knowledge*. The idea of a GA marching on a fitness landscape is maybe a little bit romantic; a suitable hill-climber is almost certainly quicker if there is a nice structure of the problem to be exploited. Maybe we should look for hard problems which can be solved somewhat slothfully by GAs but are intractable using other methods.

Another observation is that test problems for GAs were usually developed under the intuitions of the BBH. As it is believed that crossover should produce successful offspring on average, test problems were devised accordingly. So far there are no test suits that exploit the *creativity potential* of the crossover operator.

Based on these two observations a future paper is intended to present a class of problems for which GAs are hopefully well-suited.

## 7. REFERENCES

[1] J.H. Holland. *Adaptation in Natural and Artificial Systems*. University of Michigan Press, Ann Arbor, MI, 1975.

[2] D.E. Goldberg. *Genetic Algorithms in Search, Optimization &Machine Learning*. Addison Wesley, Reading, MA 1989.

[3] M. D. Vose. A critical examination of the schema theorem. Technical Report ut-cs-93-212, University of Tennessee, Computer Science Department, Knoxville, TN, USA, 1993. www.cs.utk.edu/~library/TechReports/1993/ut-cs-93-212.ps.Z.

[4] D. B. Fogel. *Evolutionary Computation: Toward a New Philosophy of Machine Learning*. IEEE Press, New York, 1995.

[5] M. Mitchell, S. Forrest, and J. H. Holland. The royal road for genetic algorithms: Fitness landscapes and GA performance. In F. J. Varela and P. Bourgine, editors, *Proceedings of the First European Conference on Artificial Life. Toward a Practice of Autonomous Systems*, pages 243-254, Cambridge, MA, 1992. MIT press.

[6] S. A. Kauffman. *The Origins of Order*, Oxford University Press, 1993.

[7] D. Whitley, K. Mathias, S. Rana, and J. Dzubera. Building Better Test Functions, *ICGA-6*, editor Eshelman, pp. 239-246, Morgan Kauffman, San Francisco, 1995.

[8] Watson, R. A., Hornby, G., and Pollack, J. B. 1998. Modeling Building-Block Interdependency. In *Proceedings of the 5th international Conference on Parallel Problem Solving From Nature,* pp. 97-108, 1998.

[9] R. A. Watson and J. B. Pollack. Symbiotic Composition and Evolvability. In *Proceedings of the 6th European Conference on Advances in Artificial Life*, pp. 480-490, 2001.

[10] M. Pelikan and D. E. Goldberg. Escaping hierarchical traps with competent genetic algorithms. *Proceedings of the Genetic and Evolutionary Computation Conference*, pp 511--518, 2001.

[11] Yu, T. and Goldberg, D. E. 2006. Conquering hierarchical difficulty by explicit chunking: substructural chromosome compression. In *Proceedings of the 8th Annual Conference on Genetic and Evolutionary Computation* (Seattle, Washington, USA, July 08 - 12, 2006). GECCO '06. ACM Press, New York

[12] R. A. Watson and J. B. Pollack. Hierarchically consistent test problems for genetic algorithms: Summary and additional results. *Late breaking papers at the Genetic and Evolutionary Computation Conference*, pp 292-297, 1999.

[13] D. Thierens and D.E. Goldberg. Mixing in genetic algorithms. *Proceedings of the Fifth International Conference on Genetic Algorithms,* pp. 38-45, 1993.

[14] M. Pelikan. *Bayesian Optimization Algorithm: From Single Level to Hierarchy*. Doctoral dissertation, University of Illinois at Urbana-Champaign, 2002.

[15] R. A. Watson. Analysis of recombinative algorithms on a non-separable buildingblock problem. *In Foundations of Genetic Algorithms VI*, pp 69-89. Morgan Kaufmann, 2001.

[16] D. H. Wolpert, W.G. Macready. No Free Lunch Theorems for Optimization, *IEEE Transactions on Evolutionary Computation*, 1, pp 1-67, 1997.

[17] Y. C. Ho, D. L Pepyne. Simple Explanation of the No-Free-Lunch Theorem and Its Implications, *Journal of Optimization Theory and Applications*, pp. 115-549, 2002.